\documentclass[letterpaper]{article} 
\usepackage{aaai24}  
\usepackage{times}  
\usepackage{helvet}  
\usepackage{courier}  
\usepackage[hyphens]{url}  
\usepackage{graphicx} 
\usepackage{natbib}  
\usepackage{caption} 
\usepackage{algorithm}
\usepackage{algorithmic}
\usepackage{svg}
\usepackage{booktabs}
\usepackage{amsmath}

\usepackage{newfloat}
\usepackage{listings}
\DeclareCaptionStyle{ruled}{labelfont=normalfont,labelsep=colon,strut=off} 
\floatstyle{ruled}
\newfloat{listing}{tb}{lst}{}
\floatname{listing}{Listing}
\title{Evolutionary Optimization of Deep Learning Agents for Sparrow Mahjong}
\author{
    Jim O'Connor,
    Derin Gezgin,
    Gary B. Parker
}
\affiliations{

    Autonomous Agent Learning Lab\\
    Connecticut College\\
    New London, Connecticut, USA\\
    \{joconno2, dgezgin, parker\}@conncoll.edu
}

\begin{document}

\maketitle

\begin{abstract}
We present Evo-Sparrow, a deep learning-based agent for AI decision-making in Sparrow Mahjong, trained by optimizing Long Short-Term Memory (LSTM) networks using Covariance Matrix Adaptation Evolution Strategy (CMA-ES). Our model evaluates board states and optimizes decision policies in a non-deterministic, partially observable game environment. Empirical analysis conducted over a significant number of simulations demonstrates that our model outperforms both random and rule-based agents, and achieves performance comparable to a Proximal Policy Optimization (PPO) baseline, indicating strong strategic play and robust policy quality. By combining deep learning with evolutionary optimization, our approach provides a computationally effective alternative to traditional reinforcement learning and gradient-based optimization methods. This research contributes to the broader field of AI game playing, demonstrating the viability of hybrid learning strategies for complex stochastic games. These findings also offer potential applications in adaptive decision-making and strategic AI development beyond Sparrow Mahjong.
\end{abstract}

\section{Introduction}

Game playing artificial intelligence (AI) has long served as a benchmark and testbed for the continued advancement of the field of AI. From early rule-based systems to modern deep reinforcement learning approaches, AI agents have demonstrated superhuman performance in various domains, including board games such as chess \cite{deepBlue}, checkers \cite{checkers}, and Go \cite{go-paper}. These advances have been driven by a combination of search-based techniques, such as minimax with alpha-beta pruning \cite{ab-pruning}, Monte Carlo Tree Search (MCTS) \cite{MCTS}, and deep learning-based policy optimization, as exemplified by AlphaGo, AlphaZero, and MuZero \cite{AlphaGo, AlphaZero, muZero}. However, while deterministic games with perfect information have been well studied, non-deterministic games with hidden information and stochastic elements continue to pose significant challenges \cite{stochastic}.

The evolution of AI for game playing has followed a trajectory of increasing complexity, beginning with early symbolic AI and rule-based systems that relied on handcrafted strategies. Early successes in this domain were driven by algorithms such as minimax, which effectively modeled decision-making in fully observable, deterministic games like chess and checkers \cite{Shannon-Chess}. The introduction of heuristic evaluations and pruning techniques, such as alpha-beta pruning, enabled AI to search deeper into game trees with reduced computational costs \cite{ab-pruning}.

The next major breakthrough in game playing AI came with probabilistic methods and machine learning-based approaches. Temporal Difference (TD) learning, exemplified by TD-Gammon \cite{TD-Gammon}, demonstrated the potential of reinforcement learning to develop sophisticated strategies without explicit domain knowledge. This system paved the way for methods that combined reinforcement learning with tree search, such as MCTS, which was instrumental in advancing AI performance in games like Go \cite{mcts-go}. MCTS's success stemmed mainly from its ability to balance exploration and exploitation efficiently, making it well-suited for large decision spaces where exhaustive search was computationally infeasible.

Deep learning further revolutionized AI game playing by enabling agents to learn directly from structured data. The introduction of deep reinforcement learning, particularly with advancements like Deep Q-Networks (DQN) \cite{DeepQLearning}, showed that neural networks could approximate value functions and learn policies for complex environments. The combination of deep learning with self-play and policy optimization, as demonstrated by AlphaGo and its successors AlphaZero and MuZero, achieved superhuman performance in Go, chess, and shogi. These advancements highlighted the power of combining deep neural networks with reinforcement learning and search techniques, setting new standards for AI in strategic decision-making.

\begin{figure}
    \centering
    \includegraphics[width=1\linewidth]{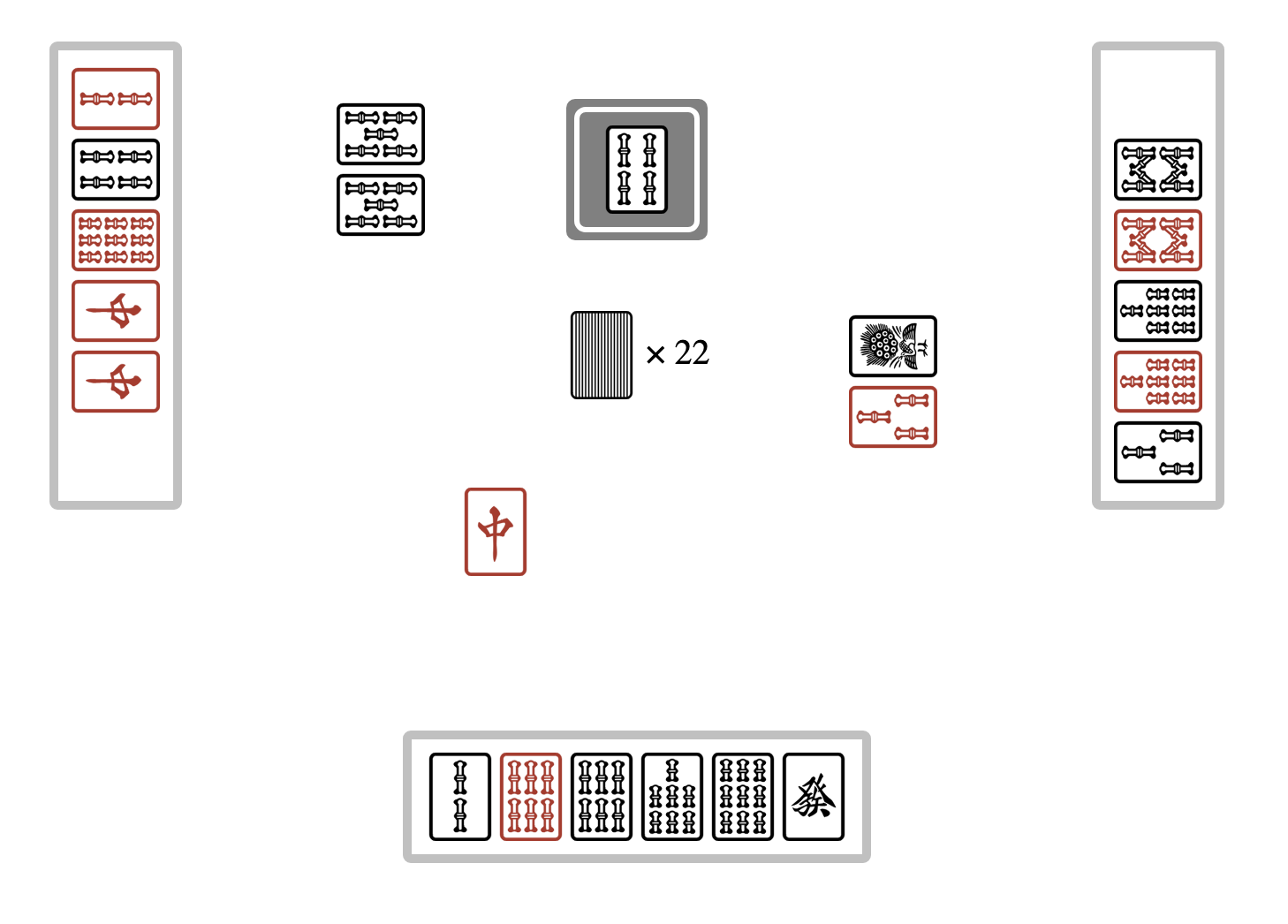}
    \caption{An example game state in Sparrow Mahjong, visualized in the PGX environment.}
    \label{fig:sample-mahjong}
\end{figure}

Mahjong, a traditional tile-based game originating in China, represents a complex domain for AI research due to its vast state space, elements of randomness, and hidden information. Variants of Mahjong have been studied within the AI community, with researchers exploring rule-based heuristics, reinforcement learning, and search-based strategies to develop competent agents \cite{suphx}. Among these variants, a simplified three-player version of the game called Sparrow Mahjong \cite{sparrow-pgx}, shown in Figure~\ref{fig:sample-mahjong}, presents an intriguing testbed for AI due to its strategic depth, partial observability, and non-deterministic nature. Unlike classic four-player Mahjong, Sparrow Mahjong reduces the number of tiles per player, accelerating the decision-making process while retaining core strategic elements.

Traditional approaches to game playing AI, such as minimax search, have limited applicability in non-deterministic, partially observable environments. Probabilistic reasoning and policy learning techniques have been employed to address these challenges in games such as poker \cite{poker} and Hanabi \cite{hanabi}, where bluffing and information inference play crucial roles. In contrast to these traditional and probabilistic methods, recent advancements in deep learning and Long Short-Term Memory (LSTM) networks \cite{LSTM} have enabled AI systems to process sequential dependencies in decision-making, making these systems well-suited for games that require long-term strategic planning.

In this work, we present Evo-Sparrow, a novel approach to training an AI agent for Sparrow Mahjong by optimizing the weights of an LSTM model using the Covariance Matrix Adaptation Evolution Strategy (CMA-ES) \cite{CMAES}. Evolutionary strategies such as CMA-ES have demonstrated effectiveness in optimizing high-dimensional neural network policies without requiring explicit gradient computation, making them a compelling alternative to traditional reinforcement learning or gradient-based optimization techniques. Our approach leverages a simple rule-based baseline to provide a structured evaluation metric, allowing us to quantify performance improvements and evaluate the efficacy of the model.

\section{Related Work}

Mahjong has been a challenging benchmark for artificial intelligence research, due to its high-dimensional state space, stochastic nature, and partial observability \cite{mahjong-partial}. According to \citet{monte-mahjong}, the state space complexity of Taiwanese Mahjong is estimated to be around $4.3 \times 10^{185}$, which exceeds the complexity of GO ($10^{172}$) and chess ($10^{46}$) \citep{complexity}. Moreover both \citet{tjong} and \citet{mahjong-competition} states that the information size of mahjong is $10^{48}$. 

Early attempts to develop a game-playing agent for Mahjong have primarily focused on rule-based systems and statistical modeling \cite{mahjong-competition}. Monte Carlo simulations and probabilistic opponent modeling have been used to predict the state of opponents, winning tiles, game scores, and upcoming moves \cite{mcts-mahjong}. While these approaches were effective in mimicking human-like strategies, they require pre-set features and struggle in generalization. One notable example of these techniques is \textsc{SIMCAT}, which utilizes a regular Monte Carlo approach with heuristic pruning for a Taiwanese variant of Mahjong. \textsc{SIMCAT} achieved the top placement in the Computer Olympiad 2020 Mahjong Tournament \cite{monte-mahjong}. 

More recent efforts in Mahjong AI have utilized deep learning. For example, a convolutional neural network (CNN) based system trained on game records showed improved discard prediction accuracy compared to the previous baselines \cite{dcnn-mahjong}. Additionally, the Suphx system \cite{suphx} represents a milestone in Mahjong AI, combining supervised learning with deep reinforcement learning and runtime policy adaptation. Suphx includes global reward prediction and oracle-guided training to handle the sparsity of rewards and the large hidden information in four-player Japanese Mahjong. Other work has explored the hierarchical decision-making to separate high-level and low-level action choices as well as the fan-backward reward allocation to handle the sparse feedback of Mahjong \cite{tjong}. Similarly, reward variance reduction techniques have been used to stabilize training in deep reinforcement learning tasks in an environment with randomness in tile draws and high variance in reward \cite{speedy-mahjong}.

All these attempts have focused on four-player mahjong variants such as Riichi Mahjong. The Sanma variant of Mahjong (three-player version of Mahjong) has additionally been addressed by Meowjong \cite{3-p-mahjong}, which uses deep CNNs pretrained with supervised learning and fine-tuned by Monte Carlo policy gradient methods. The work done through Meowjong shows that even simple variants of Mahjong keep the strategic complexity of Mahjong and still require strategic planning compared to the four-player versions. Sparrow Mahjong has not been previously studied in the field of Game AI and there are no publicly available agents or evaluation platforms for it. Our work addresses this gap by developing and evaluating a heuristic rule-based agent, a PPO-based reinforcement learning agent, and an evolutionary optimized deep learning agent, providing a comprehensive set of baselines for the Sparrow Mahjong variant.

Compared to previous approaches that utilize Monte-Carlo simulations, probabilistic modeling, and attention-based classifiers \cite{attention-mahjong}, our work takes a different approach through the combination of Long Short-Term Memory (LSTM) networks and Covariance Matrix Adaptation Evolution Strategy (CMA-ES). We find our approach to be both computationally efficient and adaptable to sparse-reward and partially observable domains such as Sparrow Mahjong. 

\section{Methodology}

\subsection{Game State Representation and Input Space}

Sparrow Mahjong is a simplified three-player variant of Mahjong that is characterized by incomplete information and stochasticity. Each player begins with 5 tiles and draws a 6th tile on their turn, after which one tile is discarded. The objective of Sparrow Mahjong is to complete a hand consisting of two melds: either sequences (e.g., 3-4-5 of the same suit) or a combination of triplets (e.g., three 7s) and one pair (two identical tiles). Notably, Sparrow Mahjong does not include some elements of traditional mahjong such as ``calling tiles''. Additionally, Dora tiles provide bonus points and are determined by a revealed indicator; all red dragons are a Dora, and one tile of each bamboo rank is designated as Dora at the start of the game.

The strategic complexity of Sparrow Mahjong arises from several key factors in the game. At certain points of the game, players can choose to draw their tiles either from the wall, analogous to a deck in other card-based games, or instead draw a tile that an opponent has just discarded. This layer of interaction makes balancing offensive versus defensive play a key component of decision-making in the game.

\begin{figure*}[!t]
    \centering
    \includegraphics[width=0.8\linewidth]{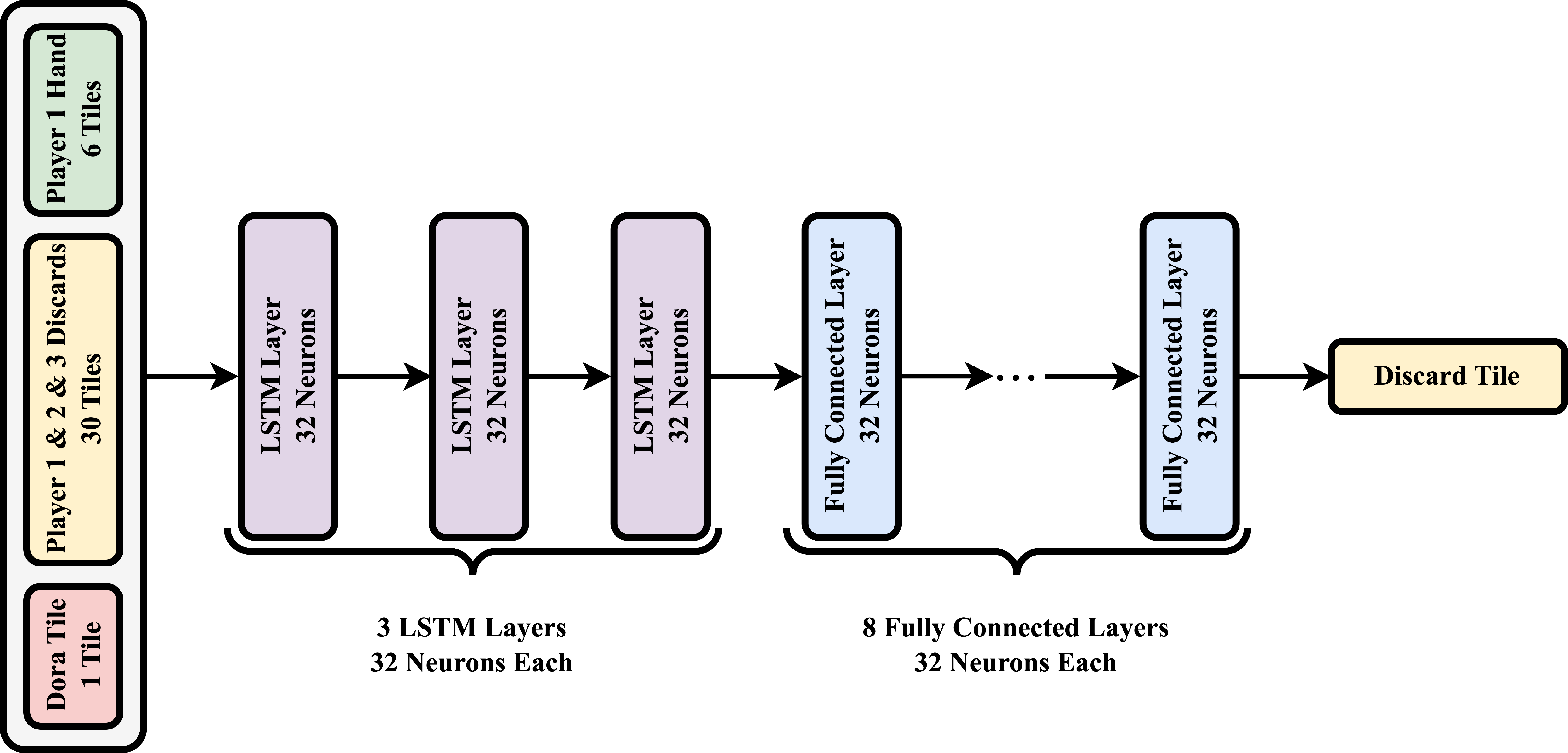}
    \caption{A high-level model of our LSTM network structure.}
    \label{fig:lstm-network}
\end{figure*}

The outcome of a Sparrow Mahjong game can be categorized into four different scenarios: win, draw, lose, and deal-in. A player wins by forming a valid hand consisting of two melds (either three consecutive numbers in the same suit or three identical tiles) and a pair of identical tiles. A draw occurs when all tiles have been drawn from the middle without any player achieving a winning hand. A player is considered to have "dealt in" when they discard a tile that directly completes an opponent's winning hand. Lastly, a player is considered to have lost when they are not responsible for the victory of the winning player. While a draw and loss do not result in a score change, a win will make the player gain points (relative to the strength of their hand), and a deal-in will make the player lose points to the winning player.

To facilitate AI learning, a selection of observable game information is numerically encoded for input into the LSTM network. The representation consists of a 37-dimensional vector capturing the player’s hand (6 integers), all players' discarded tiles (30 integers padded with 0s), and the Dora indicator (1 integer). To construct the input vector, we used the face-values as tile representations which is the default representation of the PGX \cite{sparrow-pgx} library we are using to simulate the game. This structure allows us to represent each tile by its value, and use negative values for dora tiles. This representation enables the model to distinguish between tiles without increasing the input dimensionality and infer optimal decisions based on historical game states.

\subsection{The LSTM Network}

Long Short-Term Memory Networks (LSTMs) can be used for modeling sequential decision-making tasks, especially ones that require catching long-term dependencies. Given that the game of Mahjong requires players to track a significant amount of temporal information such as past discards and opponents hands while developing long term strategies, we chose to use an LSTM-based approach for our decision network. Unlike traditional feedforward neural networks, LSTMs introduce a recurrent memory mechanism that allows the network to maintain context across multiple time steps. This gives LSTMs an advantage when addressing tasks that involve sequential dependencies, such as game playing AI, where past actions and previously seen information play a critical role in future decisions.

The core addition of LSTMs is their gated memory architecture, which is built upon three key mechanisms: the forget gate, the input gate, and the output gate. The forget gate is responsible for determining how much of the past information should be discarded at each time step. This ensures that irrelevant information does not accumulate in the network’s memory, allowing it to focus on the most relevant aspects of the game state. The input gate manages the extent to which new information is stored in the cell state. This gate is essential for selecting only the useful inputs, ensuring that the model does not become overloaded with redundant or unnecessary features. Lastly, the output gate decides what portion of the internal cell state should be passed to the next layer, which allows the model to adjust its decision-making process based on its understanding of the game environment.

The LSTM network used in this project consists of three LSTM layers followed by eight fully connected layers with 32 neurons. The LSTM layers are used for processing historical game states, and extracting important features from past moves. These features include observed opponent actions and the overall state of the game at a given point in time. By passing sequential game states through multiple LSTM layers, the network is able to encode long-term patterns that can used in the decision-making. These patterns can be processed in fully connected layers, which can generate a decision on which tile to discard. The fully connected layers act as a high-level mechanism to map learned features to actions. At each decision step, the output layer selects the tile to discard. The network structure is shown in Figure~\ref{fig:lstm-network}. The LSTM architecture was chosen via early pilot experiments with different architectures, and the final architecture was found to be the best performing in these early test. Importantly, we aimed to keep the network size manageable to ensure the efficacy of the normal variant of CMA-ES, as the covariance matrix scales quadratically with the number of parameters which can lead to excessive memory usage.

An important advantage of LSTMs over traditional Recurrent Neural Networks (RNNs) is their ability to address the vanishing gradient problem, which happens during the training of deep models. In standard RNNs, gradients tend to disappear exponentially as they are propagated through long sequences, making it difficult for the network to learn dependencies that span multiple time steps. LSTMs address this issue with their memory cell structure which maintains information across extended time steps. 

To optimize the performance of the LSTM, batch and dropout regularization were applied to prevent overfitting and enhance generalization. Batch normalization standardizes the input activations at each layer, ensuring that model the remains stable and learns efficiently. Dropout regularization prevents the model from overfitting on specific features and encourages the model for more distributed representations, leading to a better generalization across different scenarios.

\subsection{CMA-ES Optimization}

The Covariance Matrix Adaptation Evolution Strategy (CMA-ES) is a derivative-free optimization algorithm used for solving high-dimensional, non-linear optimization problems. Unlike traditional gradient-based optimization methods, CMA-ES maintains a population of candidate solutions and iteratively updates the search distribution. The optimization process models candidate solutions as samples from a multivariate normal distribution. At each iteration, CMA-ES generates a population of weight configurations, evaluates them based on gameplay performance, and updates the search distribution parameters. The global step size ($\sigma$) is dynamically adjusted via cumulative step-size adaptation (CSA) to balance exploration and exploitation. We use the standard CMA-ES variant with full covariance matrix adaptation, allowing the algorithm to model complex parameter correlations during the search. The search is initialized with a randomly sampled population and an initial step size of $\sigma = 1$. CMA-ES optimizes all $\approx 34,000$ weights of the LSTM network, where each solution represents a set of weights. Each candidate solution was evaluated 200 times against two rule-based expert agents. The cumulative score across these 200 games was recorded as the fitness value of the candidate solution. These fitness values were fed back into the CMA-ES algorithm, which updates the search distribution.

\subsection{Training and Evaluation of Evo-Sparrow}

The evaluation of each candidate solution sampled from CMA-ES was achieved by playing 200 games against two expert agents. 200 games were chosen to balance the statistical significance and keep the training time within a reasonable range. To eliminate a potential bias, the player order was randomized for each game to ensure there was no advantage or disadvantage due to the player ordering. At the end of 200 games, a fitness value was calculated for each candidate solution by summing the scores of the candidate solution for all 200 games.

After 50 generations of training, the best solution among the final set of candidate solutions is picked. To evaluate this best solution, 1,000,000 games were played against PPO-optimized and rule-based agents. 1,000,000 was chosen to reduce the variance in the evaluation and ensure statistical significance. Considering the stochastic nature of Sparrow Mahjong, a high number of trials is required to ensure statistical significance. The results from this evaluation are presented and discussed in Section~\ref{sec:results}.

\subsection{Rule-Based Expert Agent}

To benchmark performance and serve as a training opponent, a rule-based agent was developed implementing fundamental Sparrow Mahjong heuristics. The agent prioritizes tile retention based on sequence potential, avoids breaking existing sets, and minimizes discards that benefit opponents. This structured heuristic-based approach provides a robust baseline for comparison against Evo-Sparrow.

The agent’s ranking system assigns a priority value to each tile in the player’s hand. This system follows three principles. First, it prioritizes high-value tiles, as they contribute more to the player's score. Second, it avoids discarding tiles that are already part of a set or a pair. This ensures that the agent never breaks down a set that is already constructed. Lastly, the system favors keeping tiles that have a higher set-forming potential. For example, the agent avoids keeping terminal tiles (1 and 9), which have a lower potential of forming a sequence. On the other hand, it favors tiles more towards the center of the distribution which have a high potential for forming a set. Once all of the tiles are evaluated, the tile with the highest discard priority score is discarded. In cases where multiple tiles receive the same priority score, the agent selects a discard randomly from the highest-priority candidates. The agent does not perform probabilistic modeling, but follows a fixed heuristic strategy as a straightforward and simple baseline.

\subsection{Proximal Policy Optimization Agent}

To establish a strong reinforcement learning baseline, we have implemented an agent (PPO-Sparrow) that uses Proximal Policy Optimization (PPO), a widely-used and state-of-the-art policy gradient method for sequential decision-making tasks \cite{ppo-initial}. PPO has gained significant attention due to its robust performance and computational efficiency, particularly for environments characterized by partial observability and stochasticity, such as Sparrow Mahjong \cite{ppo-partial}.

PPO-Sparrow uses an LSTM-based neural network architecture that is identical to the architecture of Evo-Sparrow to ensure a fair and meaningful comparison. We made this decision to isolate the effect of the optimization method and ensure a fair comparison. The network consists of 3 LSTM layers followed by 8 fully-connected layers, each containing 32 neurons. The network outputs two distinct predictions: a policy head that generates a probability distribution over possible actions using a softmax layer, and a value head that estimates the expected cumulative reward from the current state. 

Similar to Evo-Sparrow, each training iteration consisted of 200 parallel games, enabling efficient data collection for PPO updates. Each policy update involved performing 4 epochs of optimization per collected experience batch. To ensure stable training and avoid excessive divergence between policy updates, we applied clipping to both the policy and the value-function updates. Specifically, we used a PPO clipping parameter ($\epsilon$) of 0.2, along with a learning rate of $10^{-4}$, a discount factor ($\gamma$) of 0.99, and entropy regularization to encourage exploration and prevent premature convergence. Advantage estimates were computed using Generalized Advantage Estimation (GAE) and subsequently normalized to further stabilize training. Additionally, we clipped gradients to prevent potential high-variance updates.

PPO-Sparrow was trained for $1{,}750$ generations (PPO updates) to match the exact number of games played by the Evo-Sparrow agent during training, totaling $350{,}000$ games. This setup allows for a direct comparison of learning efficiency and performance between two methods. Detailed results of PPO-Sparrow are provided in Section~\ref{sec:results}.

\section{Results}
\label{sec:results}

\subsection{Rule-Based and Random Agent Performance}
\label{sec:baseline-results}

Before evaluating Evo-Sparrow and PPO-Sparrow, we established performance benchmarks for random and rule-based agents. 

\begin{table}[t]
  \centering
  \fontsize{9pt}{10pt}\selectfont
  \setlength{\tabcolsep}{4pt}

  \caption{Performance of Random and Rule-Based agents in different match-ups over $1{,}000{,}000$ games per configuration.}
  \label{tab:rb-random-benchmarks}

  \begin{tabular}{@{}lrrrrr@{}}
    \toprule
                 & Avg.\ Score & Win \% & Draw \% & Loss \% & Deal-in \% \\
    \midrule
    Random        &  0.0023     &  8.19  & 75.52 &  5.79 & 10.50 \\
    Random        & –0.0018     &  8.15  & 75.52 &  5.82 & 10.51 \\
    Random        & –0.0006     &  8.17  & 75.52 &  5.79 & 10.52 \\
    \addlinespace
    Rule-Based    & \textbf{1.4342} & \textbf{23.83} & 61.28 & \textbf{5.38} & \textbf{9.51} \\
    Random        & –0.7120     &  7.49  & 61.28 & 12.37 & 18.87 \\
    Random        & –0.7223     &  7.49  & 61.28 & 12.27 & 18.96 \\
    \addlinespace
    Rule-Based    &  0.0045     & 19.65  & 41.52 & 15.68 & 23.15 \\
    Rule-Based    & –0.0006     & 19.63  & 41.52 & 15.70 & 23.15 \\
    Rule-Based    & –0.0039     & 19.59  & 41.52 & 15.67 & 23.22 \\
    \bottomrule
  \end{tabular}
\end{table}

Table \ref{tab:rb-random-benchmarks} presents results over 1,000,000 games, detailing average score, win rate, draw rate, loss rate, and deal-in rate. These results demonstrate the clear gap in performance between our random and rule-based agents. When all three agents were random, the average scores remained very close to zero, with nearly equal win, loss, and deal-in ratios. When a single rule-based agent was matched with two random agents, the rule-based agent significantly outperformed the random opponents with an average score of 1.4342 and a win rate of 23.83\% compared to the 7.49\% win rate of the random agents. This result confirms that our heuristic-based tile selection provides a clear advantage over a random agent and can serve as a reliable opponent for training Evo-Sparrow and PPO-Sparrow.

\subsection{Evo-Sparrow Performance}
\label{sec:cmaes-eval}

Evo-Sparrow was evaluated against both Random and Rule-Based agents across $1{,}000{,}000$ games in several mixed-agent configurations. As shown in Table~\ref{tab:dl-ppo-benchmarks}, Evo-Sparrow consistently outperforms the baseline agents across all match-ups.

In a setting with one Evo-Sparrow agent, one Rule-Based agent, and one Random agent, Evo-Sparrow achieved a win rate of 28.55\% and an average score of 0.8687, outperforming both the Rule-Based (20.91\% win rate) and Random (6.64\% win rate) agents. Evo-Sparrow also maintained a lower deal-in rate, indicating stronger defensive play alongside its offensive capability. Against two Rule-Based agents, Evo-Sparrow again led with a win rate of 26.39\% compared to the Rule-Based agents' win rates of 18.70\% and 18.71\%. Additionally, the average deal-in rate of Evo-Sparrow remained lower, showing better tile discard decisions that does not help its opponents.

\begin{table}[t]
  \centering
  \fontsize{9pt}{10pt}\selectfont
  \setlength{\tabcolsep}{4pt}

  \caption{Performance of Evo-Sparrow in mixed match-up and self-play over $1{,}000{,}000$ games per configuration.}
  \label{tab:dl-ppo-benchmarks}

  \begin{tabular}{@{}lrrrrr@{}}
    \toprule
                 & Avg.\ Score & Win \% & Draw \% & Loss \% & Deal-in \% \\
    \midrule
    \textbf{Evo-Sparrow}   & \textbf{0.8687} & \textbf{28.55} & 44.17 & \textbf{10.97} & \textbf{16.31} \\
    Rule-Based    &  0.5051         & 20.91          & 44.17 & 12.65          & 22.28 \\
    Random        & –1.3738         &  6.64          & 44.17 & 18.66          & 30.54 \\
    \addlinespace
    \textbf{Evo-Sparrow}   & \textbf{0.2648} & \textbf{26.39} & 36.65 & \textbf{14.68} & \textbf{22.28} \\
    Rule-Based    & –0.1270         & 18.70          & 36.65 & 17.15          & 27.50 \\
    Rule-Based    & –0.1378         & 18.71          & 36.65 & 17.02          & 27.62 \\
    \addlinespace
    Evo-Sparrow   & –0.0027     & 19.20 & 42.81 & 10.79 & 27.20 \\
    Evo-Sparrow  &  0.0062     & 19.25 & 42.81 & 10.77 & 27.17 \\
    Evo-Sparrow   & –0.0035     & 19.17 & 42.81 & 10.75 & 27.27 \\
    \bottomrule
  \end{tabular}
\end{table}

In order to evaluate the stability and consistency of the learned policy, we conducted a self-play evaluation using three identical instances of the final Evo-Sparrow agent. Table~\ref{tab:dl-ppo-benchmarks} shows that each agent achieved nearly identical performance in all metrics. This equivalence shows that the evolved agents do not rely on exploiting the predictable behavior of the weaker rule-based agent and demonstrate consistent performance against agents with equal strength.

\subsection{Comparative Evaluation with PPO-Sparrow}
\label{sec:ppo-baseline}

To assess the competitiveness of Evo-Sparrow against a widely used reinforcement learning baseline, we implemented PPO-Sparrow, an agent trained using Proximal Policy Optimization (PPO) with the same LSTM-based architecture as Evo-Sparrow. This comparison allows us to isolate the effects of the training methodology and directly compare the optimizers.

Table~\ref{tab:LSTM-selfplay} presents the results of a match-up where Evo-Sparrow, PPO-Sparrow, and a Rule-Based agent were evaluated over $1{,}000{,}000$ games. Evo-Sparrow achieved a slightly higher win rate (22.80\%) and average score (0.1934) compared to PPO-Sparrow (22.62\% win rate, 0.1868 score), while also maintaining a marginally lower deal-in rate (23.52\% vs.\ 23.57\%). These results` demonstrate that Evo-Sparrow is able to match or exceed the performance of a reinforcement learning based agent trained for an equal number of games. The rule-based agent in this match-up achieved the lowest win rate (19.07\%) and a significantly negative average score (–0.3802), indicating that both learned agents effectively outperform handcrafted heuristics. Statistical tests show that while the score difference between Evo-Sparrow and PPO-Sparrow is not significant based on a t-test ($p$\nobreakdash-value $= 0.1721$), a chi-square-test on win-rates confirms a statistically significant advantage for Evo-Sparrow ($p$\nobreakdash-value $= 0.0001$).

\begin{table}[!t]
  \centering
  \fontsize{9pt}{10pt}\selectfont
  \setlength{\tabcolsep}{4pt}

  \caption{Direct comparison of Evo-Sparrow, PPO-Sparrow, and Rule-Based agents in a match-up over $1{,}000{,}000$ games.}
  \label{tab:LSTM-selfplay}

  \begin{tabular}{@{}lrrrrr@{}}
    \toprule
                 & Avg.\ Score & Win \% & Draw \% & Loss \% & Deal-in \% \\
    \midrule
    \textbf{Evo-Sparrow}   & \textbf{0.1934} & \textbf{22.80} & 36.08 & \textbf{17.60} & \textbf{23.52} \\
    PPO-Sparrow   &  0.1868         & 22.62          & 36.08 & 17.74          & 23.57 \\
    Rule-Based    & –0.3802         & 19.07          & 36.08 &  9.98          & 34.87 \\
    \bottomrule
  \end{tabular}
\end{table}

In addition to performance metrics, we also compared the training efficiency of both approaches. Across 10 independent training runs, PPO-Sparrow required an average of approximately 106 minutes and 43 seconds to complete training, whereas Evo-Sparrow completed training in an average of 40 minutes and 19 seconds. This difference is primarily due to the inherent ability of CMA-ES to parallelize, which allows individuals to be distributed across multiple processes. In contrast, PPO relies on sequential policy updates and gradient-based optimization, which are inherently more difficult to parallelize. As a result, Evo-Sparrow achieves competitive performance with significantly lower training time, making it a computationally efficient alternative for learning in high-variance, partially observable environments.

\subsection{Consistency of Evo-Sparrow Training}
\label{sec:stability}

\begin{figure}[!t]
    \centering
    \includegraphics[width=\linewidth]{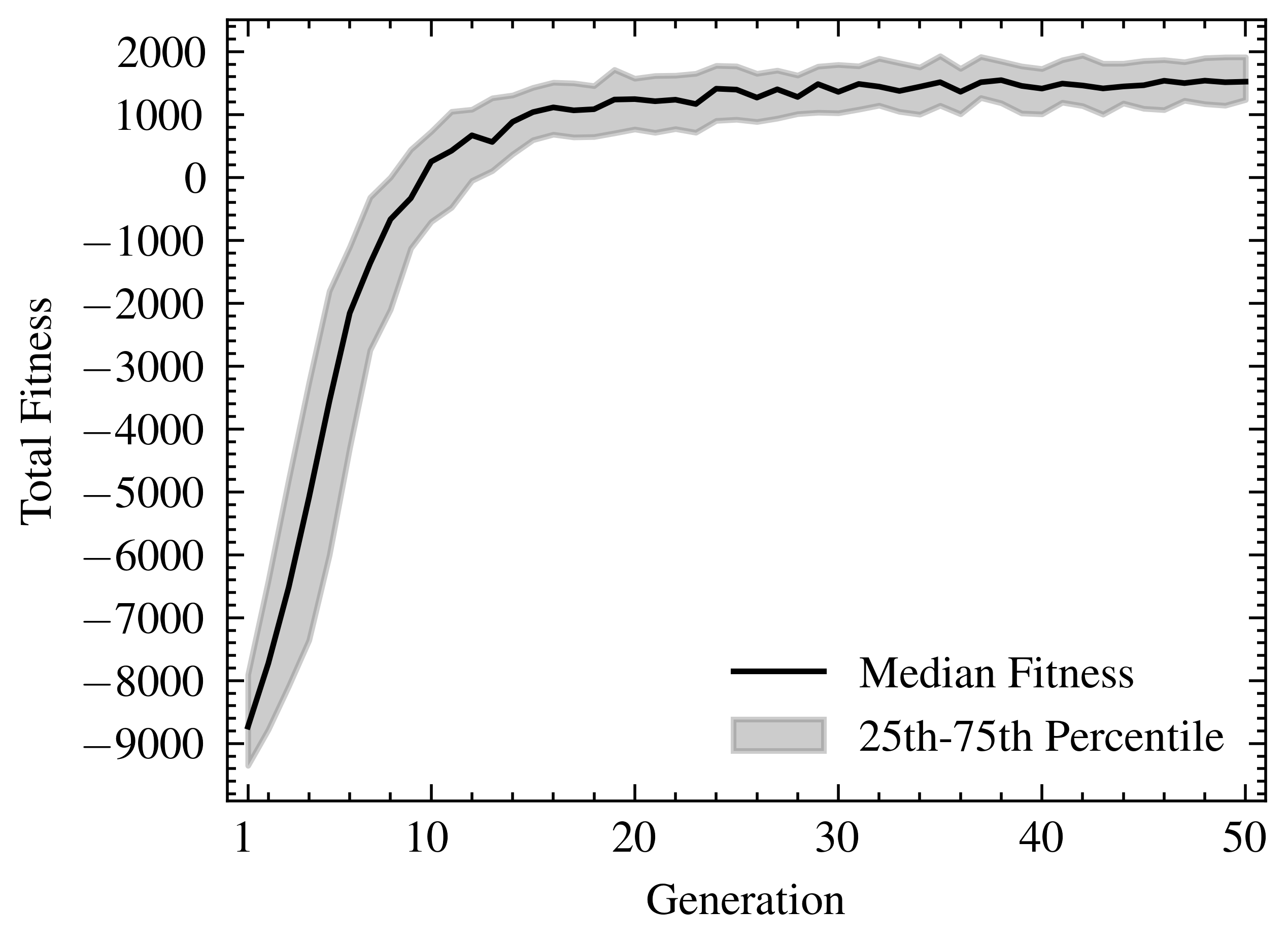}
    \caption{Median and 25th - 75th percentile fitness progression for 100 independent training runs}
    \label{fig:fitness-progression}
\end{figure}

To further validate the consistency and robustness of Evo-Sparrow, we conducted 100 independent training runs. In each run, the agent was trained for 50 generations with a population of 35 candidate solutions per generation, and every candidate solution was evaluated by playing 200 games per solution. Figure~\ref{fig:fitness-progression} illustrates the progression of total fitness across 50 generations over 100 independent training trials. The median fitness line demonstrates a fast improvement in the initial generations, climbing drastically from -9,000 at generation 1 to stable positive fitness levels by generation 20. Beyond generation 20, fitness continues to rise at a reduced rate, stabilizing near generation 40. The narrowing 25-75th percentile range shows that the variability among candidate solutions decreases as training proceeds. These results show the reliability and consistency of our approach across independent training runs.

\section{Limitations}

While our implementation shows promising results in Sparrow Mahjong, there are several limitations that constrain the generalization of this work. 

Our use of CMA-ES for optimizing the LSTM network achieves strong performance without requiring gradient information or reward shaping. CMA-ES is simpler and less GPU-intensive compared to traditional reinforcement learning methods. Nonetheless, the training still requires significant computational cost due to the need for repeated evaluations of candidate solutions across generations. Moreover, the covariance matrix updates can become memory-intensive in high-dimensional parameter spaces. Although the computation and memory costs are moderate relative to large reinforcement learning pipelines, these costs still limit the scalability to larger and more complex models.

Another limitation of this study is related to the available baselines. As Sparrow Mahjong has not been previously explored thoroughly, there are no publicly available agents or standardized benchmarks. During the training phase, we used a handcrafted rule-based agent as a baseline, which does not reflect the full capabilities of learning or search-based approaches. To address this gap, we also implemented a reinforcement learning agent using Proximal Policy Optimization (PPO), providing a stronger and more informative baseline for comparison against our evolutionary method. It remains possible to explore even more complex and larger neural architectures or to incorporate opponent modeling, hierarchical reasoning, or attention mechanisms, which may result in stronger competitors in future work.

Despite these limitations, our work demonstrates that evolutionary optimization of sequential models is an effective approach for decision-making in stochastic and partially observable games. The combination of CMA-ES and LSTM creates a robust learning framework that avoids the complexity of gradient-based methods like reinforcement learning while producing agents that outperform both random and heuristic baselines. While further enhancements are needed to generalize this approach to more complex domains, the current results provide a strong foundation for future research into hybrid optimization strategies for game AI.

\section{Conclusion}

This study demonstrates the effectiveness of combining LSTM networks with CMA-ES for decision-making in stochastic, partially observable environments. Evo-Sparrow significantly outperforms heuristic-based approaches and demonstrates competitive performance when compared to a PPO-based reinforcement learning agent. The results suggest that hybrid learning strategies can adapt well to dynamic game states, balancing offensive and defensive strategies while leveraging long-term dependencies in gameplay.

Future work can explore refinements such as incorporating opponent modeling techniques to enhance predictive accuracy and adaptability. Furthermore, expanding the architecture to handle more complex decision trees or neural architectures could further improve strategic performance. Beyond Sparrow Mahjong, the principles outlined in this study could be applied to other domains involving probabilistic decision-making in uncertain or hidden information environments. Alternative state representations, including richer encodings of tile relationships or attention-based mechanisms, can further enhance strategic decision-making. Finally, while this work focuses on Sparrow Mahjong, the proposed methodology can be extended to other stochastic, hidden-information games, and more broadly to real-world domains requiring probabilistic reasoning under uncertainty.

\bibliography{aaai24}

\begin{thebibliography}{31}
\providecommand{\natexlab}[1]{#1}

\bibitem[{Azizzadenesheli, Yue, and Anandkumar(2018)}]{ppo-partial}
Azizzadenesheli, K.; Yue, Y.; and Anandkumar, A. 2018.
\newblock Policy gradient in partially observable environments: Approximation and convergence.
\newblock \emph{arXiv preprint arXiv:1810.07900}.

\bibitem[{Bard et~al.(2020)Bard, Foerster, Chandar, Burch, Lanctot, Song, Parisotto, Dumoulin, Moitra, Hughes, Dunning, Mourad, Larochelle, Bellemare, and Bowling}]{hanabi}
Bard, N.; Foerster, J.~N.; Chandar, S.; Burch, N.; Lanctot, M.; Song, H.~F.; Parisotto, E.; Dumoulin, V.; Moitra, S.; Hughes, E.; Dunning, I.; Mourad, S.; Larochelle, H.; Bellemare, M.~G.; and Bowling, M. 2020.
\newblock The Hanabi challenge: A new frontier for AI research.
\newblock \emph{Artif. Intell.}, 280(C).

\bibitem[{Brown and Sandholm(2019)}]{poker}
Brown, N.; and Sandholm, T. 2019.
\newblock Superhuman AI for multiplayer poker.
\newblock \emph{Science}, 365: 885 -- 890.

\bibitem[{Brown, Sandholm, and Amos(2018)}]{stochastic}
Brown, N.; Sandholm, T.; and Amos, B. 2018.
\newblock Depth-limited solving for imperfect-information games.
\newblock In \emph{Proceedings of the 32nd International Conference on Neural Information Processing Systems}, NIPS'18, 7674–7685. Red Hook, NY, USA: Curran Associates Inc.

\bibitem[{Campbell, Hoane, and Hsu(2002)}]{deepBlue}
Campbell, M.; Hoane, A.~J.; and Hsu, F.-h. 2002.
\newblock Deep Blue.
\newblock \emph{Artif. Intell.}, 134(1–2): 57–83.

\bibitem[{Chen, Tang, and Wu(2022)}]{monte-mahjong}
Chen, J.; Tang, S.; and Wu, I. 2022.
\newblock Monte-Carlo Simulation for Mahjong.
\newblock \emph{Journal of Information Science and Engineering}, 38(4): 775--790.

\bibitem[{Coulom(2006)}]{MCTS}
Coulom, R. 2006.
\newblock Efficient Selectivity and Backup Operators in Monte-Carlo Tree Search.
\newblock In \emph{Computers and Games}, volume 4630, 72--83. Springer.
\newblock ISBN 978-3-540-75537-1.

\bibitem[{Gao et~al.(2019)Gao, Okuya, Kawahara, and Tsuruoka}]{dcnn-mahjong}
Gao, S.; Okuya, F.; Kawahara, Y.; and Tsuruoka, Y. 2019.
\newblock Building a Computer Mahjong Player via Deep Convolutional Neural Networks.
\newblock \emph{ArXiv}, abs/1906.02146.

\bibitem[{Gelly et~al.(2012)Gelly, Kocsis, Schoenauer, Sebag, Silver, Szepesv\'{a}ri, and Teytaud}]{mcts-go}
Gelly, S.; Kocsis, L.; Schoenauer, M.; Sebag, M.; Silver, D.; Szepesv\'{a}ri, C.; and Teytaud, O. 2012.
\newblock The grand challenge of computer Go: Monte Carlo tree search and extensions.
\newblock \emph{Commun. ACM}, 55(3): 106–113.

\bibitem[{Hochreiter and Schmidhuber(1997)}]{LSTM}
Hochreiter, S.; and Schmidhuber, J. 1997.
\newblock Long Short-Term Memory.
\newblock \emph{Neural Comput.}, 9(8): 1735–1780.

\bibitem[{Knuth and Moore(1975)}]{ab-pruning}
Knuth, D.~E.; and Moore, R.~W. 1975.
\newblock An Analysis of Alpha-Beta Pruning.
\newblock \emph{Artificial Intelligence}, 6: 293--326.

\bibitem[{Koyamada et~al.(2023)Koyamada, Okano, Nishimori, Murata, Habara, Kita, and Ishii}]{sparrow-pgx}
Koyamada, S.; Okano, S.; Nishimori, S.; Murata, Y.; Habara, K.; Kita, H.; and Ishii, S. 2023.
\newblock Pgx: Hardware-Accelerated Parallel Game Simulators for Reinforcement Learning.
\newblock In \emph{Advances in Neural Information Processing Systems}, volume~36, 45716--45743. Curran Associates, Inc.

\bibitem[{Li et~al.(2020)Li, Koyamada, Ye, Liu, Wang, Yang, Zhao, Qin, Liu, and Hon}]{suphx}
Li, J.; Koyamada, S.; Ye, Q.; Liu, G.; Wang, C.; Yang, R.; Zhao, L.; Qin, T.; Liu, T.; and Hon, H. 2020.
\newblock Suphx: Mastering Mahjong with Deep Reinforcement Learning.
\newblock \emph{CoRR}, abs/2003.13590.

\bibitem[{Li et~al.(2022)Li, Wu, Fu, Fu, Zhao, and Xing}]{speedy-mahjong}
Li, J.; Wu, S.; Fu, H.; Fu, Q.; Zhao, E.; and Xing, J. 2022.
\newblock Speedup Training Artificial Intelligence for Mahjong via Reward Variance Reduction.
\newblock In \emph{2022 IEEE Conference on Games (CoG)}, 345--352.

\bibitem[{Li et~al.(2024)Li, Liu, Wei, Wang, and Wu}]{tjong}
Li, X.; Liu, B.; Wei, Z.; Wang, Z.; and Wu, L. 2024.
\newblock Tjong: A transformer‐based Mahjong AI via hierarchical decision‐making and fan backward.
\newblock \emph{CAAI Transactions on Intelligence Technology}, 9(4): 982–995.

\bibitem[{Lu, Li, and Li(2023)}]{mahjong-competition}
Lu, Y.; Li, W.; and Li, W. 2023.
\newblock Official International Mahjong: A New Playground for AI Research.
\newblock \emph{Algorithms}, 16: 235.

\bibitem[{Mizukami and Tsuruoka(2015)}]{mcts-mahjong}
Mizukami, N.; and Tsuruoka, Y. 2015.
\newblock Building a computer Mahjong player based on Monte Carlo simulation and opponent models.
\newblock In \emph{2015 IEEE Conference on Computational Intelligence and Games (CIG)}, 275--283.

\bibitem[{Mnih et~al.(2013)Mnih, Kavukcuoglu, Silver, Graves, Antonoglou, Wierstra, and Riedmiller}]{DeepQLearning}
Mnih, V.; Kavukcuoglu, K.; Silver, D.; Graves, A.; Antonoglou, I.; Wierstra, D.; and Riedmiller, M.~A. 2013.
\newblock Playing Atari with Deep Reinforcement Learning.
\newblock \emph{CoRR}, abs/1312.5602.

\bibitem[{Ostermeier, Gawelczyk, and Hansen(1994)}]{CMAES}
Ostermeier, A.; Gawelczyk, A.; and Hansen, N. 1994.
\newblock A Derandomized Approach to Self-Adaptation of Evolution Strategies.
\newblock \emph{Evolutionary Computation}, 2(4): 369--380.

\bibitem[{Samuel(1959)}]{checkers}
Samuel, A.~L. 1959.
\newblock Some Studies in Machine Learning Using the Game of Checkers.
\newblock \emph{{IBM} Journal of Research and Development}, 3(3): 210--229.

\bibitem[{Schrittwieser et~al.(2019)Schrittwieser, Antonoglou, Hubert, Simonyan, Sifre, Schmitt, Guez, Lockhart, Hassabis, Graepel, Lillicrap, and Silver}]{muZero}
Schrittwieser, J.; Antonoglou, I.; Hubert, T.; Simonyan, K.; Sifre, L.; Schmitt, S.; Guez, A.; Lockhart, E.; Hassabis, D.; Graepel, T.; Lillicrap, T.~P.; and Silver, D. 2019.
\newblock Mastering Atari, Go, Chess and Shogi by Planning with a Learned Model.
\newblock \emph{CoRR}, abs/1911.08265.

\bibitem[{Schulman et~al.(2017)Schulman, Wolski, Dhariwal, Radford, and Klimov}]{ppo-initial}
Schulman, J.; Wolski, F.; Dhariwal, P.; Radford, A.; and Klimov, O. 2017.
\newblock Proximal policy optimization algorithms.
\newblock \emph{arXiv preprint arXiv:1707.06347}.

\bibitem[{Shannon(1950)}]{Shannon-Chess}
Shannon, C.~E. 1950.
\newblock Programming a Computer for Playing Chess.
\newblock \emph{Philosophical Magazine}, 41: 256--275.

\bibitem[{Silver et~al.(2016)Silver, Huang, Maddison, Guez, Sifre, Van Den~Driessche, Schrittwieser, Antonoglou, Panneershelvam, Lanctot et~al.}]{AlphaGo}
Silver, D.; Huang, A.; Maddison, C.~J.; Guez, A.; Sifre, L.; Van Den~Driessche, G.; Schrittwieser, J.; Antonoglou, I.; Panneershelvam, V.; Lanctot, M.; et~al. 2016.
\newblock Mastering the game of Go with deep neural networks and tree search.
\newblock \emph{Nature}, 529(7587): 484--489.

\bibitem[{Silver et~al.(2017{\natexlab{a}})Silver, Hubert, Schrittwieser, Antonoglou, Lai, Guez, Lanctot, Sifre, Kumaran, Graepel, Lillicrap, Simonyan, and Hassabis}]{AlphaZero}
Silver, D.; Hubert, T.; Schrittwieser, J.; Antonoglou, I.; Lai, M.; Guez, A.; Lanctot, M.; Sifre, L.; Kumaran, D.; Graepel, T.; Lillicrap, T.~P.; Simonyan, K.; and Hassabis, D. 2017{\natexlab{a}}.
\newblock Mastering Chess and Shogi by Self-Play with a General Reinforcement Learning Algorithm.
\newblock \emph{CoRR}, abs/1712.01815.

\bibitem[{Silver et~al.(2017{\natexlab{b}})Silver, Schrittwieser, Simonyan, Antonoglou, Huang, Guez, Hubert, Baker, Lai, Bolton, Chen, Lillicrap, Hui, Sifre, van~den Driessche, Graepel, and Hassabis}]{go-paper}
Silver, D.; Schrittwieser, J.; Simonyan, K.; Antonoglou, I.; Huang, A.; Guez, A.; Hubert, T.; Baker, L.; Lai, M.; Bolton, A.; Chen, Y.; Lillicrap, T.; Hui, F.; Sifre, L.; van~den Driessche, G.; Graepel, T.; and Hassabis, D. 2017{\natexlab{b}}.
\newblock Mastering the game of Go without human knowledge.
\newblock \emph{Nature}, 550: 354--.

\bibitem[{Tang, Chen, and Wu(2025)}]{mahjong-partial}
Tang, S.-C.; Chen, J.-C.; and Wu, I.-C. 2025.
\newblock An Efficient Method for Assessing the Strength of Mahjong Programs.
\newblock In \emph{Proceedings of the 17th International Conference on Agents and Artificial Intelligence - Volume 2: ICAART}, 124--132. INSTICC, SciTePress.
\newblock ISBN 978-989-758-737-5.

\bibitem[{Tesauro(1995)}]{TD-Gammon}
Tesauro, G. 1995.
\newblock Temporal difference learning and TD-Gammon.
\newblock \emph{Commun. ACM}, 38(3): 58–68.

\bibitem[{Truong(2021)}]{attention-mahjong}
Truong, T.-D. 2021.
\newblock \emph{A Supervised Attention-Based Multiclass Classifier for Tile Discarding in Japanese Mahjong}.
\newblock Master's thesis, University of Agder, Grimstad, Norway.

\bibitem[{{van den Herik}, Uiterwijk, and {van Rijswijck}(2002)}]{complexity}
{van den Herik}, H.; Uiterwijk, J.~W.; and {van Rijswijck}, J. 2002.
\newblock Games solved: Now and in the future.
\newblock \emph{Artificial Intelligence}.

\bibitem[{Zhao and Holden(2022)}]{3-p-mahjong}
Zhao, X.; and Holden, S. 2022.
\newblock Building a 3-Player Mahjong AI using Deep Reinforcement Learning.
\newblock \emph{ArXiv}, abs/2202.12847.

\end{thebibliography}

\end{document}